\documentclass[10pt,twocolumn,letterpaper]{article}
\usepackage{parskip} 
\usepackage{cvpr}              
\usepackage{amsmath,amssymb}
\usepackage{lmodern}
\usepackage{epsfig}
\usepackage{graphicx}
\usepackage{algorithm}
\usepackage{algorithmic}
\usepackage[algo2e,ruled,vlined]{algorithm2e}
\usepackage{setspace}
\usepackage{diagbox}
\usepackage{color}
\usepackage{booktabs}
\usepackage{float} 
\usepackage{multirow}
\usepackage[dvipsnames]{xcolor}

\usepackage[misc]{ifsym}
\definecolor{cvprblue}{rgb}{0.21,0.49,0.74}
\usepackage[pagebackref,breaklinks,colorlinks,citecolor=cvprblue]{hyperref}

\def\paperID{11193} 
\def\confName{CVPR}
\def\confYear{2024}

\def\METHODNAME{\textsc{ESA}}

\title{\METHODNAME: Energy-Based Shot Assembly Optimization \\  for Automatic Video Editing\vspace{-2mm}}

\author{Yaosen Chen$^{1,2*{~\textrm{\Letter} }}$,  Wei Wang$^{1}$,Tianheng Zheng$^{1,3}$,Xuming Wen$^{1}$,Han Yang$^{1,4}$,Yanru Zhang$^{2}$\\
		\normalsize{$^1 $Sobey Media Intelligence Laboratory, $^2 $University of Electronic Science and Technology of China, \hspace{12pt}
		}\\
		\normalsize{$^3 $SiChuan University, $^4 $Qinghai Normal University \hspace{12pt}
}\\
		{\tt\small \{chenyaosen,zhengtianheng,wangwei,wenxuming,yanghan\}@sobey.com  yanruzhang@uestc.edu.cn}\hspace{12pt}\\
}

\renewcommand*{\thefootnote}{\fnsymbol{footnote}}
\begin{document}
	


\twocolumn[{%
		\renewcommand\twocolumn[1][]{#1}%
		\maketitle
	}] 
	

\footnotetext{$*$ Equal Contribution. $~\textrm{\Letter}$ Corresponding Author.}

\begin{abstract}
Shot assembly is a crucial step in film production and video editing, involving the sequencing and arrangement of shots to construct a narrative, convey information, or evoke emotions. Traditionally, this process has been manually executed by experienced editors. While current intelligent video editing technologies can handle some automated video editing tasks, they often fail to capture the creator’s unique artistic expression in shot assembly. To address this challenge, we propose an energy-based optimization method for video shot assembly. Specifically, we first perform visual-semantic matching between the script generated by a large language model and a video library to obtain subsets of candidate shots aligned with the script semantics. Next, we segment and label the shots from reference videos, extracting attributes such as shot size, camera motion, and semantics. We then employ energy-based models to learn from these attributes, scoring candidate shot sequences based on their alignment with reference styles. Finally, we achieve shot assembly optimization by combining multiple syntax rules, producing videos that align with the assembly style of the reference videos. Our method not only automates the arrangement and combination of independent shots according to specific logic, narrative requirements, or artistic styles but also learns the assembly style of reference videos, creating a coherent visual sequence or holistic visual expression. With our system, even users with no prior video editing experience can create visually compelling videos. Project page: \href{https://sobeymil.github.io/esa.com/}{https://sobeymil.github.io/esa.com/}.
\end{abstract}

\renewcommand{\thefootnote}{\arabic{footnote}}
\section{Introduction}
\label{Introduction}
With advancements in technology, intelligent tools designed to assist users with limited experience in creative processes are becoming increasingly abundant. These tools are widely used in image editing, drawing, and even 3D modeling and manufacturing. However, for beginners, the creation and editing of videos remain a challenging process. Professional video editors typically use editing software such as Adobe Premiere\footnote{https://www.adobe.com/products/premiere.html} or Sobey Editmax\footnote{https://www.sobey.com/productsDetails?index=1\&idx=9} to process raw footage and generate coherent videos based on narratives or storylines. In contrast, non-professionals often lack the cinematic or aesthetic knowledge required for video editing and may find these software applications difficult to operate. This high barrier restricts more users from engaging in video creation, thereby hindering the widespread dissemination of creative expression.

\begin{figure}[]
		\centering
		\includegraphics[width=1\columnwidth]{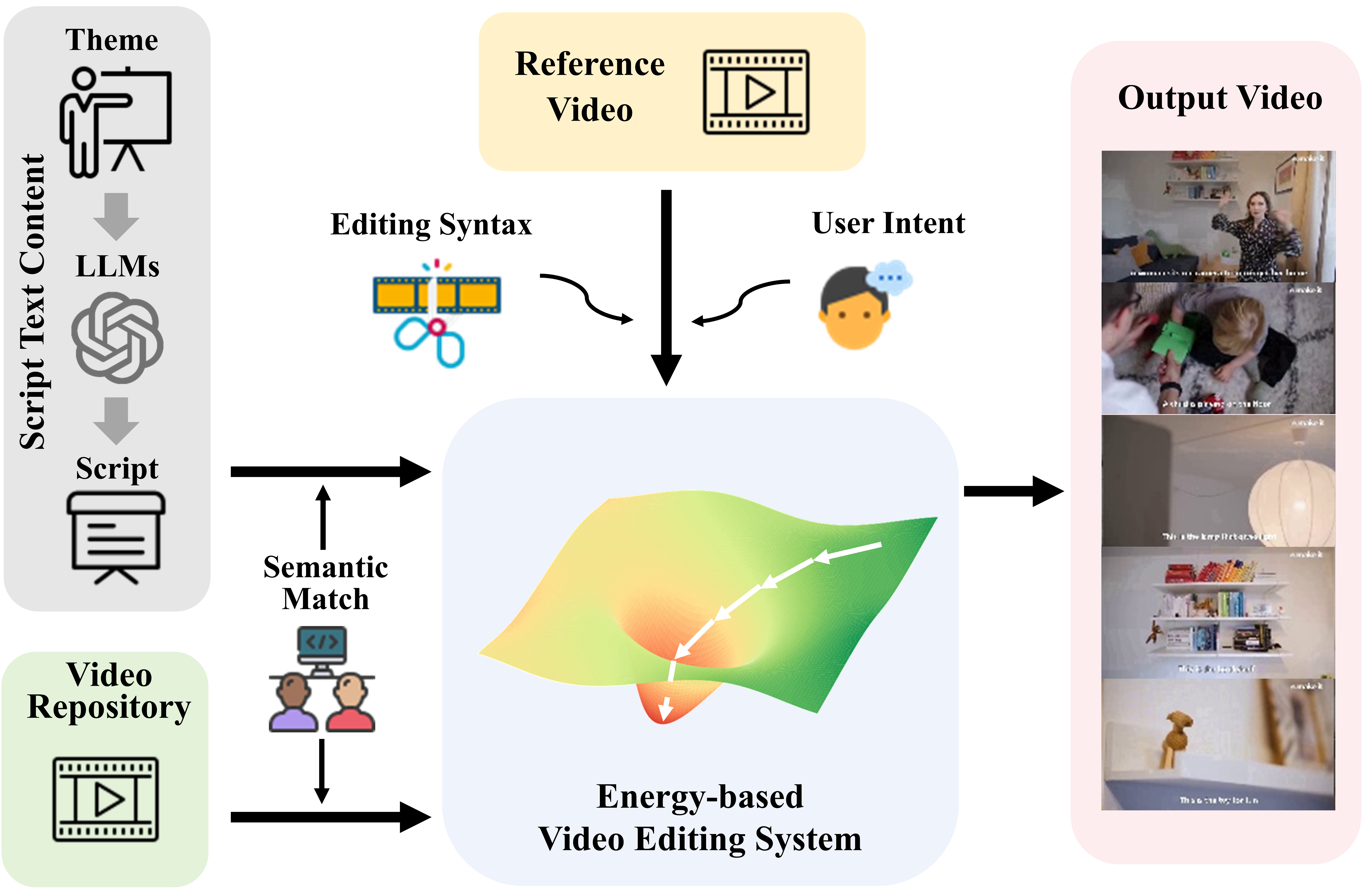}
		\caption{\textbf{Energy-based shot assembly optimization.} Given a specific theme and a library of video clips, our method employs an energy-based model to search for an optimal shot assembly that aligns with the thematic semantics, editing syntax, and user intent. Combined with additional post-production processes such as voice-overs and subtitles, this approach enables high-quality, intelligent video editing.}
		\label{fig:results_overview}
	\end{figure}

In film and video editing, constructing a coherent and engaging narrative structure relies on the process of shot assembly. By carefully selecting, ordering, and juxtaposing shots, experienced editors can convey information, establish pacing, and evoke desired emotional responses. Commonly, this complex process has heavily depended on human expertise and intuition. Skilled editors leverage their experience, creativity, and an implicit understanding of cinematic ``Syntax" to ensure that shot selection and arrangement produce a narrative style consistent with the director’s vision or the conventions of a particular genre. Despite the growing sophistication of intelligent video editing tools, replicating these nuanced artistic considerations within automated systems remains a significant challenge.

Recently, advancements in artificial intelligence have led to great progress in automated video editing. State-of-the-art methods can segment clips, recognize key actions, detect semantic content, and even generate preliminary edits based on predefined themes. However, while these tools excel at basic tasks—such as trimming, concatenating, or rearranging clips—they often fail to capture the subtle narrative cues, visual continuity, and expressive rhythms that characterize human-driven shot assembly. Thus, there remains a noticeable gap between automated editing technologies and the complex, stylistically consistent outputs produced by skilled human editors.

The primary reason for this gap lies in the intricate interplay of aesthetic, semantic, and syntactic factors that guide shot assembly. Effective shot assembly is not merely about simple content matching or visual similarity; it requires balancing narrative logic, continuity rules, stylistic preferences, and contextual semantics. Editors internalize these ``syntax rules" through experience and reference to established cinematic traditions, enabling them to craft sequences that are both purposeful and artistic. To emulate this process, computational approaches must not only recognize the semantic alignment between scripts and available video resources but also learn how reference shots are structured and combined to generate a unique style.

Addressing these challenges, this study proposes an energy-based optimization framework for automated video shot assembly, aiming to bridge the gap between current AI-driven editing solutions and the nuanced artistry of human editorial practices. As shown in Figure \ref{fig:results_overview}, our approach integrates three key components. First, we employ large language models to generate scripts related to the video repository based on the given theme, and further match candidate shots with semantic alignment to scripts from the video repository. Second, we segment reference videos and annotate their shots with attributes such as shot size, motion, and semantic content. These attributes provide the foundation for modeling the ``syntax" of shot assembly—the rules and patterns that guide how individual shots are combined into a coherent whole. Finally, we employ energy-based models to learn from this syntax, scoring candidate shot sequences based on their alignment with reference styles. By optimizing these learned energy functions, our system can automatically assemble shots in a manner that respects narrative logic, pacing, and the style derived from human-edited reference materials. Our contributions are threefold:
\begin{itemize}[leftmargin=0.1cm, itemindent=0.1cm]
	\item First, we introduce a method that combines semantic analysis of scripts, shot attributes, and reference-driven syntax modeling to achieve style-consistent shot assembly. 
	\item Second, we demonstrate how the energy-based optimization framework can score and optimize shot arrangements based on learned representations of stylistic and narrative rules.
	\item Third, we show that our approach significantly lowers the barrier to high-quality video editing, enabling users without professional editing experience to create visually and narratively coherent videos that reflect a target aesthetic.
\end{itemize}

In the following sections, we review related work in intelligent video editing, energy-based modeling, and semantic-visual alignment(Section \ref{related_work}). We then defines the problem of automated video editing by optimizing shot size transition scores and introduces Langevin sampling as a method to solve this discrete optimization problem(Section \ref{Preliminaries}).

We then detail our system’s methodology(Section \ref{methods}), describe the experimental setup, and present results that highlight the system’s ability to replicate stylistic shot assembly patterns(Section \ref{experiments}). Finally, we discuss limitations, potential applications, and avenues for future research(Section \ref{sec:Limitations}). Through this work, we take an important step toward more intelligent, artistically informed video editing tools, enabling more non-professional users to easily engage in video creation and realize their creative visions.

\section{Related work}
\label{related_work}

\noindent\textbf{ Intelligent Video Editing. }
Post-production involves video editing, color correction~\cite{chen2023nlut,jiang2024ncst}, sound processing, and adding visual effects to integrate raw footage into a complete film using professional editing software. Advances in artificial intelligence have significantly improved the efficiency and quality of video processing. For example, methods like NLUT~\cite{chen2023nlut} achieve color correction comparable to professional colorists but with greater efficiency. Additionally, temporal action detection~\cite{chen2021boundary,chen2021capsule} and video summarization~\cite{chen2022video} techniques enhance the creation of highlight reels for sports events and variety shows.

Recent studies have leveraged generative AI for intelligent video editing. MVOC~\cite{wang2024mvoc} composites multiple independent video objects, while TVGs~\cite{zhang2024tvg} enables natural transitions between video segments. Hyperlips~\cite{chen2023hyperlips} modifies speakers’ lips in videos to maintain audio-visual synchronization. VCoME~\cite{gong2024vcome} generates coherent and visually appealing verbal videos by integrating multimodal editing effects across text, visual, and audio categories. Despite these advancements, existing methods do not fully address the entire video production workflow from a system-level perspective.

The Cinematographic Camera~\cite{jiang2024cinematographic} Diffusion Model is a Transformer-based diffusion model that manages rhythm and utilizes the stochastic nature of diffusion models to generate diverse, high-quality camera trajectories based on high-level textual descriptions, adhering to video shooting standards. However, it is limited to video generation and does not support editing pre-recorded videos. Similarly, Write-A-Video~\cite{wang2019write} allows users to create video montages through text editing by compiling clips from online repositories or personal albums, but it cannot assemble shots to match a specific artist’s proprietary editing style.

\noindent\textbf{Energy-based Models.}
In recent years, Energy-Based Models (EBMs)~\cite{hinton2002training,song2021train} have demonstrated remarkable representational and optimization capabilities in various visual and multimodal tasks. Unlike discriminative or generative models that directly predict target outcomes, EBMs employ an energy function to evaluate the quality of a given configuration, making them well-suited for complex, multi-constraint scenarios. During training and inference, EBMs typically leverage Markov Chain Monte Carlo (MCMC) and approximate Langevin dynamics to explore the energy landscape~\cite{song2021train}. By introducing stochastic perturbations and energy-guided updates, these methods efficiently search high-dimensional, combinatorial spaces to find low-energy (high-quality) solutions, thus laying a solid foundation for constructing and optimizing layered, compositional energy landscapes.

\noindent\textbf{Semantic-Visual Alignment.}
In the realm of text-to-image generation, Park et al.~\cite{park2024energy} incorporate EBMs within cross-attention layers to iteratively update the posterior of context vectors, significantly improving semantic alignment between generated images and input text. Zhang et al.~\cite{zhang2025object} further introduce object-conditioned EBM constraints to refine attention maps, ensuring more accurate attribute binding and object-specific semantics in the generated output. Meanwhile, in scene rearrangement and planning tasks, Gkanatsios et al.~\cite{gkanatsios2023energy} employ EBMs to map linguistic descriptions to spatial energy predicates. By performing gradient descent over multiple predicate energies, they achieve object configurations that satisfy multiple instructions simultaneously, without additional training for complex combinations. Additionally, ``Write-A-Video”~\cite{wang2019write} utilizes segment-level energy functions to measure the semantic, rhythmic, and stylistic alignment between video segments and scripts, employing dynamic programming for global optimization. Building on these advances, our work applies EBMs to video shot assembly. We integrate visual-semantic matching, assembly rules derived from reference videos, and aesthetic considerations into a unified energy framework. This enables automated, multi-constraint optimization of shot sequences that are both logically coherent and stylistically consistent, making high-quality video editing accessible to users without professional experience.

\section{Preliminaries}
\label{Preliminaries}
\subsection{Problem Definition}

In video production and post-production, video editing must adhere to certain syntactic characteristics to convey information more effectively visually, guide audience emotions, and enhance the viewing experience. These syntactic characteristics, often referred to as visual language or editing rules, leverage the combination of shot size, camera motion, and semantics to serve narrative and emotional expression. Automatic editing that conforms to these syntactic rules is accomplished through algorithmic analysis and optimization, which automatically selects and arranges video segments to comply with professional guidelines for shot size, camera motion, visual language, and narrative logic.

\begin{table}[ht]
	\centering
	\setlength{\tabcolsep}{4pt}
	\caption{An example of the shot size transition score matrices: It quantifies transition scores between consecutive shot sizes.}
	\label{tab:sc}
	\small 
		\begin{tabular}{c|c|c|c|c|c}
			\toprule
			\diagbox[width=8em]{Prev Shot}{Next Shot} & ELS & LS &MS & CU & ECU \\ 
			\midrule
			ELS & 0.0 & 0.5 & 1.0 & 0.0 & 0.0 \\ \hline
			LS & 0.5 & 0.6 & 1.0 & 1.0 & 0.0 \\ \hline
			MS & 1.0 & 1.0 & 1.0 & 1.0 & 1.0 \\ \hline
			CU & 0.0 & 0.6 & 0.8 & 0.6 & 1.0 \\ \hline
			ECU & 0.0 & 0.0 & 0.3 & 1.0 & 0.0 \\ 
			\bottomrule
		\end{tabular}
	\vspace{-2mm}
\end{table}

Given a set of video clips \(\mathbf{V} = \{ v_1, v_2, \dots, v_N \}\), where \( N \) is the number of video clips, we need to select \( K \) clips and arrange them into a sequence \(\mathbf{S} = (s_{1}, s_{2}, \dots, s_{K})\) that maximizes the \textbf{total shot size syntax score}:
\begin{align}
	\label{eq:Score}
	\text{Score}(\mathbf{S}) = \sum_{i=1}^{K-1} G_{t_{s_{i}}, t_{s_{i+1}}}
\end{align}
Here, \( G_{t_{s_{i}}, t_{s_{i+1}}} \) represents the transition score between consecutive shot sizes.

Define the \textbf{set of shot sizes} \(\mathbf{T} = \{ t_1, t_2, \dots, t_M \}\), where \( M \) is the number of shot size categories. In this paper, we assume \( M = 5 \). A \textbf{shot size mapping} associates each video clip \( v_i \) with a shot size \( t_{v_i} \in \mathbf{T} \). The \textbf{shot size syntax score matrix} \(\mathbf{G}\in \mathbb{R}^{M\times M}\) defines the transition scores \(G_{ij}\) from shot size \(t_{i}\) to shot size \(t_j\).
Table \ref{tab:sc} shows an example of the shot size syntax score matrix.
Extreme Long Shot (ELS): depicts a very distant viewpoint, often providing an overview of the entire scene. Long Shot (LS): clearly shows the whole subject (e.g., a full-body view of a person) along with some environment. Medium Shot (MS): displays part of the subject, commonly from the waist up, often used in dialogue scenes. Close-Up (CU): focuses on a specific detail of the subject (e.g., face, hands) to highlight emotions or details. Extreme Close-Up (ECU): further magnifies details such as eyes, fingers, or other small objects to convey intense emotion or subtle actions.

\subsection{Langevin Sampling}

We define a \textbf{selection matrix} \(\mathbf{X}\in \{0,1\}^{N\times K}\), where \(X_{ik}=1\) indicates that the video clip \(v_i\) is chosen at position \( k \); otherwise \(X_{ik}=0\). Based on the definitions above, the selection matrix must satisfy the following constraints:

\begin{align}
	\label{eq:sinkhorn1}
	\sum_{i=1}^{N} X_{ik} = 1, \quad \forall k \in \{1, 2, \dots, K\}   \\[6pt]
	\sum_{k=1}^{K} X_{ik} \leq 1, \quad \forall i \in \{1, 2, \dots, N\} 
\end{align}

The first constraint enforces that ``each position selects exactly one video clip”, and the second ensures that ``each video clip is selected at most once”.

Our goal is to minimize the following energy function:

\begin{align}
	\label{eq:eg}
	E(\mathbf{X}) = -\sum_{k=1}^{K-1} \sum_{i=1}^{N} \sum_{j=1}^{N} X_{ik} X_{j,k+1} G_{t_{s_i}, t_{s_j}},
\end{align}

that is:

\begin{align}
	\mathbf{X}^* = \arg \min_{\mathbf{X}} E(\mathbf{X}),
\end{align}

to obtain the optimal \textbf{selection matrix} \(\mathbf{X}^*\).

To optimize and find the optimal \(\mathbf{X}^*\), we can naturally consider leveraging Langevin dynamics-based sampling. However, Langevin sampling optimization operates in continuous space, meaning \(\mathbf{X}\) must be a continuous variable. To transform the discrete variable (the binary selection matrix \(\mathbf{X}\)) into a continuous representation, a common method involves normalizing the selection matrix with the Sinkhorn algorithm \cite{cuturi2013sinkhorn}. This normalization not only converts the discrete variables into continuous values but also ensures that the constraints defined in Eq. \ref{eq:sinkhorn1} are satisfied. Algorithm S4 in \textit{Supplementary Material} shows the process of obtaining the continuous matrix \(\mathbf{X}_{\text{s}}\). This enables the application of Langevin dynamics-based sampling to optimize the continuous variable \(\mathbf{X}_{\text{s}}\). When a binary solution is required, a reversion to \(\mathbf{X}_{\text{h}}\) is possible.

Thus, the Langevin dynamics-based optimization of the continuous selection matrix \(\mathbf{X}_{\text{s}}\) is:

\begin{align}
	\mathbf{X}^{t+1}_{\text{s}} &= \mathbf{X}^t_{\text{s}} - \eta \nabla_\mathbf{X} E(\mathbf{X}^t_{\text{s}}) + \sqrt{2 \eta \epsilon} \mathbf{\xi}^t,
	\label{eq:Langevin}
\end{align}

where \(\mathbf{\xi}^t \sim \mathbf{N}(0, I)\) is a Gaussian noise term, \(\eta\) is the learning rate, and \(\epsilon\) denotes the noise intensity.

\begin{figure*}[t!]
	\centering
	\includegraphics[width=1.0\textwidth]{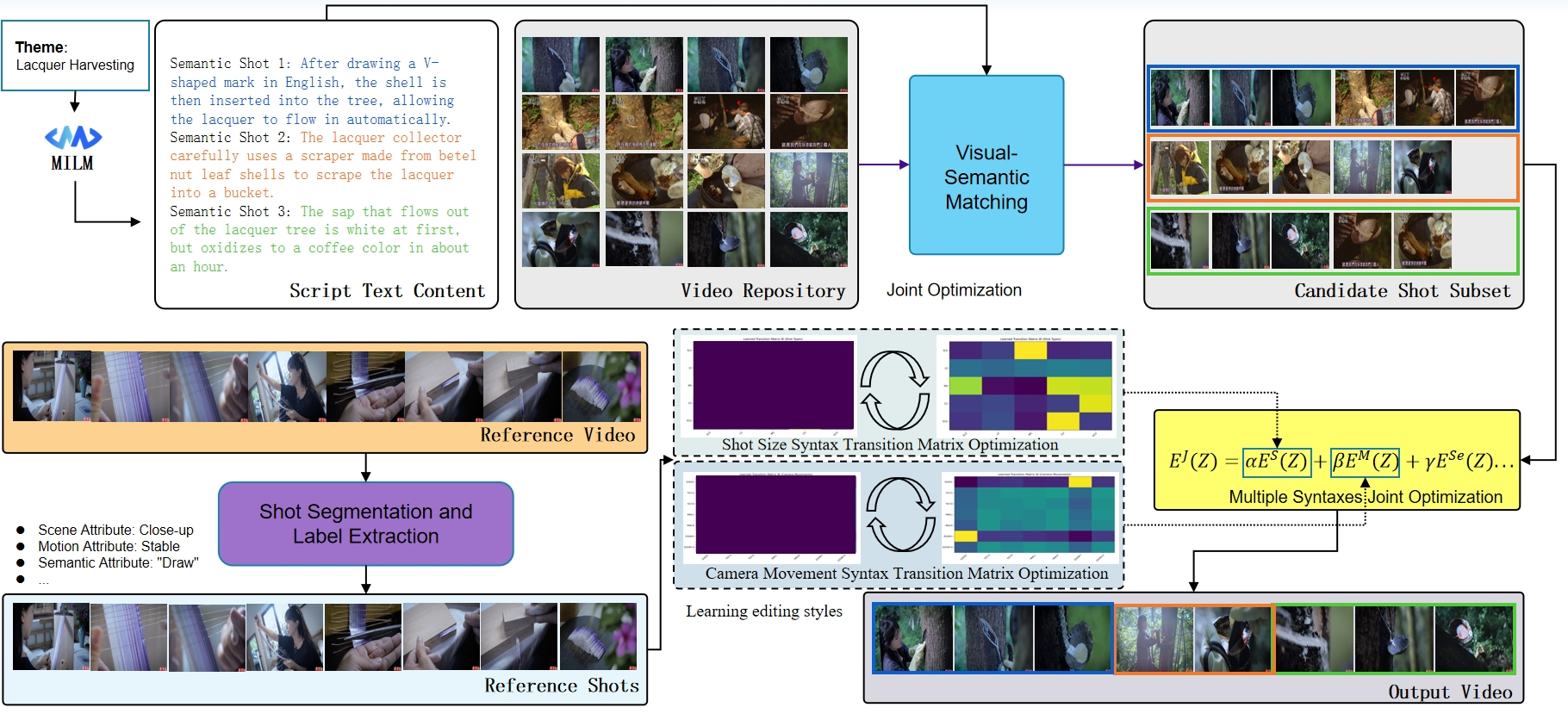}
	\caption{\textbf{Overall Framework of Energy-Based Shot Assembly Optimization for Automatic Video Editing.} Our approach consists of modules such as ``Shot Segmentation and Label Extraction", ``Visual-Semantic Matching", and ``Multiple Syntaxes Joint Assembly Optimization". These modules work together to automatically generate an edited video sequence that reflects the shot scale style and camera motion style of a reference sequence.}
	\label{fig:mvoc_framework}
\end{figure*}

\section{Our Approach}
\label{methods}
\subsection{Overall Framework}

As shown in Figure \ref{fig:mvoc_framework}, our framework takes a textual description as input and generates a sequence of video shots. The core idea is to integrate semantic analysis, visual matching, and energy-based optimization into a multimodal joint optimization process.

Given a user-provided textual description, we first retrieve candidate shots from a video library that semantically match the text. By visual-semantic matching, we filter out an initial set of candidate shots. We then segment these candidate shots and extract shot-related labels. These labels, combined with shot syntax rules, provide structured semantic guidance for the shot assembly optimization. Building on this foundation, the framework employs an energy-based model to evaluate the quality of the combined sequence. The energy model considers multiple objectives, including the alignment of textual semantics with video content, visual continuity between shots, temporal logic consistency, and adherence to shot syntax rules.

\subsection{Assembly Optimization}

We aim to select a fixed number of video shots from a given set and arrange them into a sequence that maximizes the sum of transition scores between consecutive shots. This is a classic combinatorial optimization problem, which requires identifying the highest-scoring sequence from a vast number of possible permutations.

Two commonly used methods in combinatorial optimization are Beam Search (BS) and Genetic Algorithms (GA). BS is a heuristic search technique that performs local neighborhood operations—such as swaps or replacements—and retains the top-B partial solutions to guide the search without exhaustively enumerating all candidates. GA, on the other hand, iteratively evolve a population of solutions by applying selection, crossover, and mutation operations to progressively obtain better solutions. \textit{For details on BS Assembly Optimization and GA Optimization algorithms, please refer to the Supplementary Material.}

\subsection{Langevin-like Updates for Assembly Optimization}\label{subsec:multipleDependence}

For discrete combinatorial optimization, continuous Langevin sampling requires transforming discrete variables into continuous space and relies on gradient information. This limits its direct ability to find global optima in combinatorial spaces. While Beam Search can handle discrete problems effectively, it may struggle in global search due to limited beam width. Genetic Algorithms provide global search capabilities but lack refined local optimization strategies, resulting in potentially slower convergence.

Inspired by both continuous Langevin sampling and combinatorial optimization heuristics, we propose a \textbf{Langevin-like update method} that directly optimizes the discrete sequence $\mathbf{S}$, rather than its continuous relaxation. Integrating Langevin-like updates into either a pure Beam Search or pure GA strategy can provide additional local search capabilities and controlled randomness, enabling the algorithm to better balance global exploration and local improvement. This, in turn, can help escape local optima, speed up convergence, and enhance overall solution quality.

We previously define continuous Langevin sampling for a selection matrix \(\mathbf{X}_{s} \in \mathbb{R}^{N\times K}\) in Eq. \ref{eq:Langevin}. To avoid confusion, let us reconsider the Langevin dynamics equation. In continuous space, the Langevin equation describes particle motion in a potential field, combining gradient descent with random noise:
\begin{align}
	\mathbf{Z}^{t+1} &= \mathbf{Z}^{t} - \eta \nabla_{\mathbf{Z}} E(\mathbf{Z}^{t}) + \sqrt{2 \eta \epsilon} \mathbf{\xi}^{t}
	\label{eq:LangevinZ}
\end{align}
where $\mathbf{Z}^{t}$ is the state at iteration $t$, $\eta$ is the learning rate, $\epsilon$ is the noise intensity (temperature), and $\mathbf{\xi}^{t} \sim \mathbf{N}(0,I)$ is Gaussian noise.

In discrete space, states $\mathbf{Z}$ cannot be differentiated to obtain $\nabla E(\mathbf{Z})$. Instead, we approximate the gradient step by energy differences between discrete states. The continuous gradient step $-\eta \nabla_{\mathbf{Z}} E(\mathbf{Z}^{t})$ is replaced by evaluating $\Delta E = E(\mathbf{Z}') - E(\mathbf{Z}^{t})$ for a candidate new state $\mathbf{Z}'$ from the neighborhood. Drawing inspiration from \cite{chib1995understanding}, we define a transition probability to decide whether to accept $\mathbf{Z}'$:
\begin{align}
	P_{\text{accept}} = \min\left(1, \exp\left(-\frac{\Delta E}{\epsilon}\right)\right)
\end{align}

Thus, the discrete Langevin-like update rule is:
\begin{align}
	\label{eq:DiscreteL}
	\mathbf{Z}^{t+1} =
	\begin{cases}
		\mathbf{Z}', & \text{if } \Delta E \leq 0 \text{ or } r < P_{\text{accept}} \\[4pt]
		\mathbf{Z}^{t}, & \text{otherwise}
	\end{cases}
\end{align}
where $r$ is a random number in $[0,1]$.

In our context, $\mathbf{Z}$ corresponds to the discrete shot sequence $\mathbf{S} = (s_1, s_2, \dots, s_K)$. To improve the performance of optimization methods such as Beam Search and Genetic Algorithms, we introduce a discrete Langevin-like update rule, as described in the local optimization step of Algorithm \ref{alg:DiscreteLangevinSampling}. This update enhances the search process by incorporating local refinement and controlled randomness, allowing the acceptance of improved solutions within a neighborhood or suboptimal solutions with a probability derived from energy differences. \textit{For details on Langevin-like updates for beam
search assembly optimization, please refer to the Supplementary Material.}

Building on this idea, we develop a hybrid optimization strategy that integrates the local optimization capability of Langevin-like sampling with the global search advantage of Genetic Algorithms. The proposed approach, outlined in Algorithm \ref{alg:DiscreteLangevinSampling}, employs Langevin-like updates for local exploration and combines them with genetic operations—selection, crossover, and mutation—to maintain population diversity. Through iterative evolution, the method effectively balances intensification and diversification, gradually converging toward high-quality solutions.

\begin{algorithm}[]
	\caption{Assembly Optimization with Langevin-like Updates and GA (Langevin+GA)}
	\label{alg:DiscreteLangevinSampling}
	\footnotesize
	\KwIn{
		A set of video shots $\mathbf{V}=\{v_1,\ldots,v_N\}$;
		Sequence length $M$; Maximum number of iterations $I$;
		Number of desired top solutions $Q$;
		Energy function $E$;
		Langevin parameters: step size $\epsilon$, temperature $T$;
		GA parameters: crossover probability $r_c$, mutation probability $r_m$, population size $P$.
	}
	
	\KwOut{
		Top sequences and their scores.
	}
	
	\BlankLine
	\textbf{Initialize}: Randomly generate $P$ individuals $\mathcal{P} = \{\mathbf{S}_1, \ldots, \mathbf{S}_P\}$; set $E_{\text{best}} \leftarrow \infty$.\\	
	\For{$iter = 1$ to $I$}{
		\tcp{Langevin-like Local Optimization}
		$\mathcal{P}' \leftarrow \emptyset$; \\
		\ForEach{$S \in \mathcal{P}$}{
			Compute $E(S)$;\\
			Generate the neighborhood $\mathcal{N}(S)$;\\
			Find $S_{\text{best\_neighbor}} = \arg\min_{S' \in \mathcal{N}(S)} E(S')$;\\
			Compute $\Delta E = E(S_{\text{best\_neighbor}})-E(S)$; \\
			Compute $P_{\text{accept}} = \min(1,\exp(-\Delta E / (\epsilon T)))$; \\
			\If{$\Delta E < 0$ or random number $< P_{\text{accept}}$}{
				$S \leftarrow S_{\text{best\_neighbor}}$
			}
			Insert the updated $S$ into $\mathcal{P}'$.
		}
		
		\tcp{Genetic Algorithm Operations}
		Compute fitness: $F(\mathbf{S}) = -E(\mathbf{S})$ for $\mathbf{S} \in \mathcal{P}'$;\\
		Compute selection probabilities:
		\[
		p(\mathbf{S}) = \frac{\exp(F(\mathbf{S}))}{\sum_{\mathbf{S}' \in \mathcal{P}'} \exp(F(\mathbf{S}'))}.
		\]
		Sample $P$ sequences from $\mathcal{P}'$ according to $p(\mathbf{S})$ to form a parent candidate set $\mathcal{P}''$;\\
		$\mathcal{P}_{new}=\emptyset$;\\
		\While{$|\mathcal{P}_{new}|<P$}{
			Randomly select two parents $S_{p1}, S_{p2}\in\mathcal{P}''$;\\
			With probability $r_c$, perform crossover to produce offspring $(C_1,C_2)$; otherwise $(C_1,C_2)\leftarrow (S_{p1},S_{p2})$;\\
			Mutate $C_1$ and $C_2$ with probability $r_m$;\\
			Insert offspring into $\mathcal{P}_{new}$.
		}
		$\mathcal{P}\leftarrow\mathcal{P}_{new}$;\\
		\If{\text{enough high-quality solutions ($>Q$) are found}}{
			\textbf{break}
		}
	}
	
	\Return{Top sequences and their corresponding scores.}
	
\end{algorithm}

\subsection{Predicting Shot Size Syntax Styles}
The shot assembly optimization methods described so far assume a known syntax score matrix (e.g., as in Table \ref{tab:sc}). In practical scenarios, however, we may wish to emulate the editing style from a given reference video without explicitly knowing the shot size syntax rules beforehand. In other words, we aim to learn a scoring function that evaluates how closely a candidate sequence aligns with the shot size syntax style of a given reference sequence.

Consider a reference sequence $\mathbf{R} = [r_1, r_2, \ldots, r_M]$ of shot types extracted from professionally edited videos. We derive a parametric transition matrix $W_{\text{ref}}$ that captures the statistical patterns of shot sequencing in $\mathbf{R}$ through the following procedure:

\begin{enumerate}
	\item \textbf{Categorical Mapping}:  
	Each shot type $r_i \in \mathcal{S}$ in $\mathbf{R}$ is mapped to an index $k \in \{0,1,\dots,K-1\}$ via a bijection $\phi: \mathcal{S} \to \{0,1,\dots,K-1\}$, where $\mathcal{S} = \{\text{ELS, LS, MS, CU, ECU}\}$ and $K = |\mathcal{S}| = 5$.
	
	\item \textbf{Transition Counting}:  
	We construct an empirical count matrix $C \in \mathbb{N}^{K \times K}$ where each entry:  
	\begin{equation}
		C_{ij} = \sum_{t=1}^{M-1} \mathbb{I}\left(\phi(r_t) = i \land \phi(r_{t+1}) = j\right)
	\end{equation}
	counts transitions from shot size $i$ to $j$ in $\mathbf{R}$, with $\mathbb{I}$ denoting the indicator function.
	
	\item \textbf{Probability Estimation}:  
	The reference transition matrix $W_{\text{ref}}$ is obtained by normalizing $C$:  
	\begin{equation}
		W_{\text{ref}} = \frac{C}{S}, \quad S = \sum_{i=0}^{K-1} \sum_{j=0}^{K-1} C_{ij}
	\end{equation}
	where $S$ represents the total number of observed transitions. This yields a row-stochastic matrix where $W_{\text{ref}}[i,j]$ estimates the transition probability $P(\text{shot}_{t+1} = j \mid \text{shot}_t = i)$.
\end{enumerate}

This statistically derived $W_{\text{ref}}$ serves as a prior distribution encoding professional editing conventions, providing the foundation for evaluating syntactic coherence in candidate sequences $\mathbf{C}$. The matrix captures both dominant transition patterns (e.g., MS $\to$ CU) and rare transitions through its probability mass distribution, effectively modeling shot sequencing as a first-order Markov process.

\begin{table}[h!]
	\centering
	\setlength{\tabcolsep}{4pt}
	\resizebox{\columnwidth}{!}{
		\begin{tabular}{c|c|c|c|c|c|c|c}
			\toprule
			\diagbox{Prev Shot}{Next Shot} & Stable & Up & Down & Left & Right & Out & In \\ \midrule
			Stable & 1.0 & 1.0 & 1.0 & 1.0 & 1.0& 1.0& 1.0 \\ \hline
			Up & 1.0 & 1.0 & 0.0 & 0.0 & 0.0& 1.0& 1.0 \\ \hline
			Down & 1.0 & 0.0 & 1.0 & 0.0 & 0.0& 1.0& 1.0 \\ \hline
			Left & 1.0 & 0.0 & 0.0 & 1.0 & 0.0& 1.0& 1.0 \\ \hline
			Right & 1.0 & 0.0 & 0.0 & 0.0 & 1.0& 1.0& 1.0 \\ \hline
			Out & 1.0 & 0.0 & 0.0 & 0.0 & 0.0& 1.0& 0.0 \\ \hline
			In & 1.0 & 0.0 & 0.0 & 0.0 & 0.0& 0.0& 1.0 \\ \bottomrule
		\end{tabular}
	}
	\caption{An example of the motion transition score matrices: It quantifies transition scores between consecutive motion attributes.}
	\label{tab:mv}
\end{table}

\subsection{Multiple Syntaxes Joint Optimization}\label{subsec:VideoObjectDependence1}
We have focused on shot size syntax optimization thus far. In reality, a shot may have multiple attributes, such as camera motion. In these scenarios, the final optimization should jointly consider multiple syntactic dimensions. For example, consider both shot size syntax, motion syntax, and semantic syntax.

Similar to the shot size syntax definition, we define the \textbf{set of motion types} \(\mathbf{m} = \{ m_1, m_2, \dots, m_L \}\), where \( L \) is the number of motions categories. In this paper, we assume \( L = 6 \). Table \ref{tab:mv} shows an example of the motion syntax score matrix. Then, we define a \textbf{motion syntax score matrix} \( \mathbf{M}\in \mathbb{R}^{L\times L}\), where \(M_{m_im_j}\) indicates the transition score from motion type \(m_i\) to \(m_j\). Maximizing the total motion syntax score:
\begin{align}
	\label{eq:mvScore}
	\text{Score}(\mathbf{M}) = \sum_{i=1}^{K-1} M_{m_{s_{i}}, m_{s_{i+1}}}
\end{align}
is equivalent to minimizing a motion energy function:
\begin{align}
	\label{eq:emv}
	E^M(\mathbf{Z}) = -\sum_{k=1}^{K-1} \sum_{i=1}^{N} \sum_{j=1}^{N} X_{ik} X_{j,k+1} M_{m_{s_i}, m_{s_j}}.
\end{align}

To further ensure the accuracy of semantic matching, a semantic energy function is defined. Specifically, for a given discrete shot sequence $\mathbf{S}$ (representing state $\mathbf{Z}$), each shot possesses a predefined text description $\mathbf{D} = (d_1, d_2, \dots, d_K)$. This description $\mathbf{D}$ and the input Script Text Content $\mathbf{T}$ are encoded into vectors using the CLIP text encoder $EN_{text}$. The cosine similarity between these two vectors then measures the degree of semantic matching between the discrete shot sequence and the Script Text Content. Consequently, the objective is to minimize this semantic energy function:

\begin{align}
	\label{eq:es}
	E^{Se}(\mathbf{Z}) = -\text{cos\_sim}(EN_{text}(\mathbf{D}), EN_{text}(\mathbf{T})).
\end{align}

Let the shot size syntax energy be \(E^G(\mathbf{Z})\) as defined in Eq. \ref{eq:eg}, and let \(E^M(\mathbf{Z})\) and \(E^{Se}(\mathbf{Z})\) be the energy models for motion and semantic syntax, respectively. Joint optimization seeks to minimize:
\begin{align}
	\label{eq:emv1}
	E^J(\mathbf{Z}) = \alpha E^G(\mathbf{Z}) + \beta E^M(\mathbf{Z}) + \gamma E^{Se}(\mathbf{Z})
\end{align}
where \(\alpha, \beta, \gamma\) are coefficients that balance the different components. Replacing \(E(\mathbf{Z})\) in Eq. \ref{eq:LangevinZ} with \(E^J(\mathbf{Z})\) allows Langevin-like sampling to perform joint optimization over multiple syntactic attributes.

\begin{figure*}[t!]
	\centering
	\includegraphics[width=0.95\textwidth]{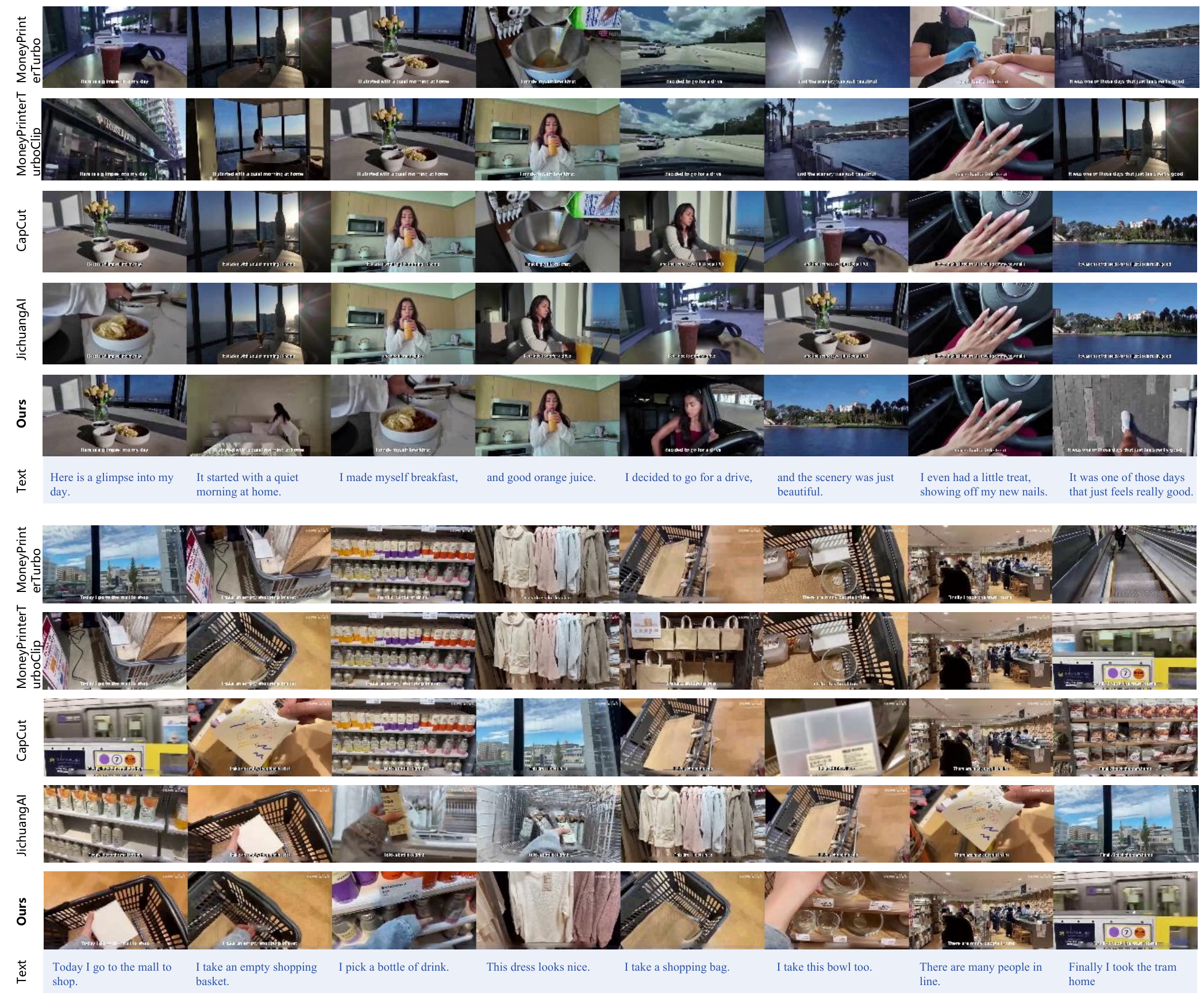}
	\caption{\textbf{Visual Comparison of Video Editing.} We compare the results of video editing using the same ``Script Text Content" and ``Video Repository", and our method achieves better visual-text alignment.}
	\label{fig:VideoEditingVisualization}
\end{figure*}

\section{Experiments}
\label{experiments}
\subsection{Experimental Setup}
In the experiments, six candidate video repositories, each containing a variety of shot sizes, are constructed. Shots are selected from these repositories for assembly. The default experimental parameters are set as follows: the energy model is trained for 100 steps, and the maximum number of Langevin sampling steps is 1000.

For the hardware configuration, all experiments are conducted on a server equipped with an NVIDIA GeForce RTX 3090 GPU to ensure consistency in computational resources. This GPU features 24GB of video memory, which is sufficient to meet the demands of large-scale data processing and model training. The algorithm’s runtime on the CPU is recorded to provide a comprehensive evaluation of its efficiency.

To ensure the reliability of the results, each method undergoes multiple tests, and the average optimization accuracy and resource usage are calculated. It is noted that the training time is influenced by factors such as video length. By adjusting parameters, the training time can vary within a range of 20 to 30 minutes.

\subsection{Evaluation}

\begin{table*}
	\centering
	\caption{Comparison of Subjective Video Similarity Scores. This table presents the performance of different video generation methods in experiments based on two distinct reference videos. The methods are evaluated on four metrics: Semantic Matching Score (SMC), Camera Motion Similarity (CMS), Shot Size Similarity (SSS), and Overall Style Similarity (OSS). These metrics collectively reflect the stylistic similarity between the generated and reference videos.}
	\label{tab:similarity_scores}
	\begin{tabular}{@{}l|cccccccc@{}}
		\toprule
		\multirow{2}{*}{Method} & \multicolumn{4}{c}{Reference1} & \multicolumn{4}{c}{Reference2}   \\
		
		\cmidrule(lr){2-5} \cmidrule(lr){6-9} 
		& SMC   & CMS   & SSS   & OSS& SMC   & CMS   & SSS   & OSS   \\ 
		\midrule
		MoneyPrinterTurbo       & 3.190  & 2.571  & 2.048  &  2.190  & 1.619  & 2.167  & 2.286  &  1.905  \\
		MoneyPrinterTurboClip   & 3.381  & 2.905    & 2.190    & 2.381& 2.619  & 2.810  & 2.667  &  2.571 \\ 
		CapCut                  &   2.333  & 2.286    & 1.952   & 1.952& 2.810  & 2.643  & 2.857  &  2.667 \\ 
		JichuangAI              &   2.524  & 2.429    & 1.952   & 2.000 & 1.714  & 2.738  & 2.619  &  2.429\\ 
		\textbf{Ours}           &  \textbf{3.524}  & \textbf{3.095} &  \textbf{2.476 }  & \textbf{2.619} &  \textbf{3.286}  & \textbf{3.381} &  \textbf{3.095 }  & \textbf{3.095}    \\
		\bottomrule
	\end{tabular}
\end{table*}

\begin{table*}
	\centering
	\caption{Comparison of Objective Video Similarity Scores. This table presents the Mean Squared Error (MSE) between the transition matrices of the generated videos and those of the reference video. Specifically, it compares the Shot Size and Camera Motion transition matrices. These metrics provide an objective assessment of the stylistic similarity between the generated and reference videos. Here, M\_MSE and S\_MSE denote the MSE for camera motion and shot size transition matrices, respectively.}
	\label{tab:Objective_scores}
	\begin{tabular}{@{}l|cccc@{}}
		\toprule
		\multirow{2}{*}{Method} & \multicolumn{2}{c}{Refernce1} & \multicolumn{2}{c}{Refernce2}   \\
		
		\cmidrule(lr){2-3} \cmidrule(lr){4-5} 
		& M\_MSE   & S\_MSE   & M\_MSE   & S\_MSE   \\ 
		\midrule
		MoneyPrinterTurbo       & 0.059  & 0.174  & 0.066  &  0.101  \\
		MoneyPrinterTurboClip   & 0.061  & 0.201    & 0.037    & 0.087 \\ 
		CapCut                  &   0.064  & 0.201    & 0.031   & 0.105 \\ 
		JichuangAI              &   0.055  & 0.121    & 0.029   & 0.123 \\ 
		\textbf{Ours}           &  \textbf{0.048}  & \textbf{0.055} &  \textbf{0.021 }  & \textbf{0.079}     \\
		\bottomrule
	\end{tabular}
\end{table*}
\begin{figure*}[t!]
	\centering
	\includegraphics[width=0.95\textwidth]{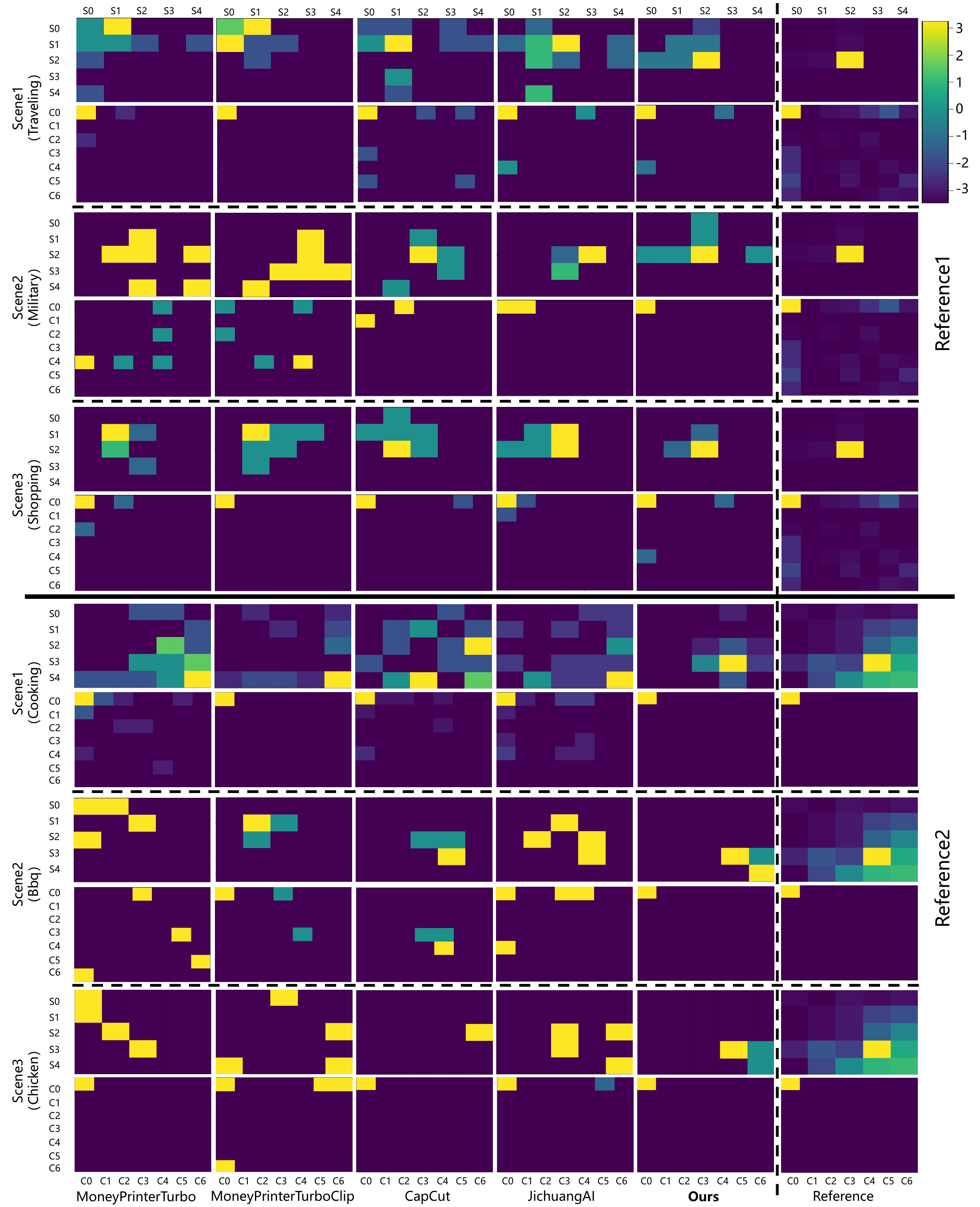}
	\caption{\textbf{Visual Comparison of the Transition Score Matrices for Shot Size and Camera Motion Syntax in the Edited Videos.} S0 to S4 respectively represent the shot size attributes: Extreme Long Shot (ELS), Long Shot (LS), Medium Shot (MS), Close-Up (CU), and Extreme Close-Up (ECU). C0 to C6 respectively represent the camera motion attributes: Stable, Up, Down, Left, Right, Out, and In.}
	\label{fig:TransitionMatrixVisualization}
\end{figure*}

The experiments focus on evaluating the performance of Langevin sampling optimization based on the energy model and the style similarity of shot concatenation. We primarily use \textbf{Editing Style Similarity Assessment} to evaluate the performance of the assembled video. This assessment measures the style consistency between the assembled video and the reference video using both subjective and objective metrics.

\subsubsection{Experimental Setup.}
In the editing style similarity assessment experiment, two reference videos are selected for the experiment. Reference video 1 is an excerpt from the TV series \textit{iPartment}, while reference video 2 is an excerpt from a \textit{Li Ziqi} video. The primary motion style of reference video 1 is the stable shot, with the main shot size being the medium shot. Reference video 2 also primarily uses stable shots, but its main shot size is the close-up shot. For reference video 1, three material libraries of varying scales are constructed: traveling, military, and shopping, containing 121, 62, and 50 shots respectively. For reference video 2, three material libraries are also constructed: cooking, bbq, and chicken, with 224, 36, and 50 shots respectively.

We compare our method against four video editing tools: (1) \textbf{MoneyPrinterTurbo} \footnote{https://github.com/harry0703/MoneyPrinterTurbo}, an open-source automated short video editing tool based on large language models. Its core functionality involves coordinating multiple modules (e.g., script generation, speech synthesis, material matching, video composition) through a task orchestration system, enabling an automated ``input-theme-to-output-video" pipeline. (2) However, its local upload scheme fails to achieve alignment between textual descriptions and visual content. Therefore, we integrate AltCLIP \cite{altclip} into this framework to optimize text-visual alignment, naming our modified version \textbf{MoneyPrinterTurboClip}. (3) \textbf{CapCut} \footnote{https://www.capcut.com/zh-tw}'s ``Smart Templates" feature allows users to select preset templates to automatically generate complete videos from uploaded images or video clips. It supports personalized adjustments such as text overlays and music, with direct export capability for sharing high-definition outputs to social media platforms. (4) \textbf{JichuangAI} \footnote{https://aic.oceanengine.com/} provides a template-based video generation feature similar to CapCut’s. It integrates functionalities such as AI-powered script generation, video creation, digital human synthesis, and voice-over production.

\subsubsection{Comparison of Subjective Video Similarity Scores} 
We present a visual comparison of video editing results in Fig.~\ref{fig:VideoEditingVisualization}. Under the same ``Script Text Content" and ``Video Repository" as the input, our method demonstrates superior performance in visual-text alignment.

To evaluate style similarity, we assemble a panel of expert assessors who score videos generated by different methods according to four standardized criteria outlined in Table~\ref{tab:similarity_scores}: \textbf{Semantic Matching Score (SMC)}, measures alignment between video captions/subtitles and visual content; \textbf{Camera Motion Similarity (CMS)}, evaluates similarity to the reference video based exclusively on transition patterns of camera motion (e.g., zooms, pans, tilts, dolly shots); \textbf{Shot Size Similarity (SSS)}, evaluates similarity to the reference video based exclusively on transition patterns of shot scales (e.g., long shots, close-ups, medium shots); \textbf{Overall Style Similarity (OSS)}, comprehensively assesses similarity to the reference video in overall editing style, incorporating camera motion transitions, shot size transitions, and caption matching (i.e., SMC).

All criteria are rated on a standardized scale from 0 to 5. We invite 30 volunteers to conduct anonymous evaluations of the video editing results. As shown in Table~\ref{tab:similarity_scores}, our method outperforms competing approaches across all assessed dimensions.

\subsubsection{Comparison of Objective Video Similarity Scores}

In our experiments, we primarily use shot sizes, camera motion, and semantics as the syntactic rules for video shot assembly. Since semantics are related to the corresponding ``Script Text Content", we mainly focused on examining the similarity in shot size and camera motion styles between the assembled videos and the reference videos. We visualize the transition score matrices of shot sizes and camera motion in the reference videos and compare them with those of videos assembled using various methods, as shown in Fig. \ref{fig:TransitionMatrixVisualization}. Then, based on these visualized values, we calculate the corresponding Mean Squared Error (MSE) for comparison, as presented in Table \ref{tab:Objective_scores}. In  Fig. \ref{fig:TransitionMatrixVisualization}, the peaks of the transition matrices obtained by our method align almost perfectly with those of the reference videos across all test cases, while other methods show significant discrepancies. This indicates that our method effectively learns the stylistic rules of shot sizes and camera motion from the reference videos. Similarly, in Table \ref{tab:Objective_scores}, the MSE values we calculated are the smallest among all methods, further demonstrating the effectiveness of our approach.

\begin{table*}[h!]
	\centering
	\caption{Theoretical analysis of the characteristics and complexity of four methods.}
	\label{tab:comparison}
	\resizebox{\textwidth}{!}{
		\begin{tabular}{lcccc}
			\toprule
			Characteristics & Langevin+GA  & Langevin+BS  & GA-only  & BS-only  \\ 
			\midrule
			Global Search Mechanism & Genetic (selection/crossover/mutation) & Beam retains multiple paths & Genetic Algorithm & Beam retains multiple paths \\[6pt]
			Local Search Optimization & Yes (Langevin) & Yes (Langevin) & No & No \\[6pt]
			Source of Diversity & Genetic + random neighborhood & Beam multi-solutions + neighborhood search & Random crossover and mutation & Top $B$ solutions and expand neighborhoods \\[6pt]
			Escape from Local Optima & High (Genetic + acceptance) & Moderate (Langevin acceptance) & High (mutation) & Low (greedy retention) \\[6pt]
			Parameter Complexity & High (many parameters) & Moderate (Beam+Langevin) & Low (pure GA) & Low (pure Beam) \\[6pt]
			Complexity & $O(I P K M)$ & $O(I B K M)$ & $O(I P M)$ & $O(I B K M)$ \\ 
			\bottomrule
		\end{tabular}
	}
	
	\vspace{0.5em}
	\footnotesize{
		\textbf{Parameter Descriptions:}\\
		\textbf{\(I\)}: max number of iterations; \quad
		\textbf{\(P\)}: population size (for GA); \quad
		\textbf{\(B\)}: beam size (for Beam Search); \quad
		\textbf{\(K\)}: neighborhood size; \quad
		\textbf{\(M\)}: sequence length.
	}
\end{table*}

\begin{table*}
	\centering
	\caption{Optimization accuracy test results. ``5 choose 3 random" indicates that candidate videos do not have all aspect labels, with a total of 200 test samples; ``Extensively random" indicates that the number of candidate videos includes 5, 6, 7, 8, 9, 10, and the number of selected videos includes 3, 4, 5, 6, 7, with a total of 240 test samples.}
	\label{tab:com}
		\begin{tabular}{c|c|c}
			\hline
			Methods				 & 5 choose 3 random & Extensively random\\ \hline
			Continuous Langevin & 81.50\% & 55.00\%  \\ 
			\hline
			BS-only & \textbf{100.00\%} & 94.17\% \\ 
			GA-only  & \textbf{99.50\%} & 93.75\% \\ 
			Langevin + BS & \textbf{100.00\%} & 98.75\% 
			\\ 
			Langevin + GA   & \textbf{100.00\%} & \textbf{100.00}\% \\
			\hline
		\end{tabular}
	\vspace{-2mm}
\end{table*}

\subsubsection{Ablation Study}
This ablation study is designed to validate the effectiveness of the core optimization strategy within our proposed method. Instead of evaluating all modules, we focus on a critical question: is the hybrid discrete-continuous optimization strategy necessary and superior to alternative approaches? To answer this, we systematically compare the performance of our proposed hybrid algorithms against standalone baselines under varying problem complexities.

\paragraph{Experimental Setup for Algorithm Comparison}
To ensure a comprehensive evaluation, we define three testing scenarios with increasing complexity:
\begin{itemize}
	\item \textbf{Fixed 5-choose-3}: A constrained scenario where candidate videos contain all necessary shot labels (60 test samples).
	\item \textbf{Random 5-choose-3}: A more challenging scenario where candidate videos may lack some shot labels (200 test samples).
	\item \textbf{Extended random}: The most complex and realistic scenario, with 5-10 candidate videos and 3-7 videos to select (240 test samples).
\end{itemize}
The primary metric is \textbf{optimization accuracy}—the success rate of finding the globally optimal sequence. We compare five methods:
\begin{itemize}
	\item \textbf{Continuous Langevin}: A baseline applying Langevin sampling directly to a continuous relaxation of the problem.
	\item \textbf{BS-only}: Beam Search only (Algorithm 1), a classic discrete search.
	\item \textbf{GA-only}: Genetic Algorithm only (Algorithm 2), a population-based discrete search.
	\item \textbf{Langevin+BS}: Hybrid of Langevin (local) and Beam Search (global) (Algorithm 3).
	\item \textbf{Langevin+GA}: Hybrid of Langevin (local) and Genetic Algorithm (global) (Algorithm 4).
\end{itemize}
\paragraph{The Necessity of a Hybrid Optimization Strategy}
The central question is whether combining local Langevin updates with a global search algorithm provides a measurable benefit. The results in Table \ref{tab:com} provide a clear affirmative answer. In the most complex extended random scenario, the hybrid methods \textbf{Langevin+GA} and \textbf{Langevin+BS} achieved near-perfect accuracies of 100\% and 98.75\%, respectively. This performance significantly surpasses that of the standalone global searches, \textbf{GA-only} (93.75\%) and \textbf{BS-only} (94.17\%). This demonstrates that the Langevin component's ability to perform local refinement is a critical factor for reliably finding the optimal sequence in complex search spaces, justifying the hybrid design.

\paragraph{Performance Scaling with Problem Complexity}
A robust algorithm must maintain performance as problem difficulty increases. The data reveals a clear trend: while all discrete optimization methods (BS-only, GA-only, and the hybrids) perform well (accuracy $>$ 99.5\%) in the simpler 5-choose-3 scenarios, a significant gap emerges in the extended random scenario. The accuracy drop for the standalone algorithms is more pronounced than for the hybrids. This indicates that the advantage of the hybrid strategy is not merely incremental but becomes essential for solving harder, more realistic problems. The hybrid approach's ability to balance global exploration (via GA or BS) with local exploitation (via Langevin) makes it uniquely scalable.

\paragraph{The Inapplicability of Purely Continuous Optimization}
We also investigate whether the discrete nature of the problem can be bypassed using a continuous optimization method. The results decisively show it cannot. The \textbf{Continuous Langevin} baseline achieved the lowest accuracy in all scenarios (81.50\% in 5-choose-3 and 55.00\% in extended random). This stark performance gap underscores that the discrete sequence optimization problem cannot be effectively solved by treating it as a continuous task. This finding validates our core design decision to develop a method that explicitly operates in the discrete domain, using Langevin dynamics as a component within a discrete framework rather than as the primary optimizer.

Based on the systematic analysis above, the selection of \textbf{Langevin+GA} as the final optimization strategy is justified by empirical evidence. It consistently delivers top-tier performance across all scenarios, with its superiority being most critical in the most complex cases. The genetic algorithm's population-based approach provides a natural and effective mechanism for global exploration, which is synergistically enhanced by the local refinement capability of Langevin sampling. This ablation study conclusively validates that the proposed hybrid strategy is a key contributor to the method's overall effectiveness, proving its necessity and superiority over simpler algorithmic alternatives.

\section{Limitations}
\label{sec:Limitations}
Although the energy model-based video editing optimization method proposed in this paper achieves significant progress in automated video editing, it still has several limitations that require further investigation in future research.  First, the Langevin-style update process is computationally intensive, limiting its efficiency for long videos or large datasets. Future work could explore approximate inference or model simplification to improve speed.
Second, while the model adapts well to a single reference style, it cannot effectively blend editing syntaxes from multiple reference videos—a useful ability for combining diverse artistic styles. Enabling such composition would need enhanced mechanisms to represent and reconcile multiple style priors.

\section{Conclusion}
\label{sec:Conclusion}
This work proposes a video editing optimization method based on energy model, aiming to address the lack of artistry and stylistic consistency in automated video editing. Through semantic analysis, visual matching, and energy model optimization, the method achieves automatic selection and arrangement of video shots, learns the editing style of a reference video, and generates high-quality videos that align with user intent. Experimental results demonstrate that the method outperforms existing video editing tools in terms of both optimization accuracy and style similarity, proving its effectiveness and superiority. The introduction of this method provides a new direction for the development of intelligent video editing technology, helping to lower the barrier to video creation and enabling more people to participate in video production and creative expression.

{
	\small
	\bibliographystyle{unsrt}
	\bibliography{main}
}
	

\clearpage
\section*{Supplementary Material}
\renewcommand\thesection{\Alph{section}}
\setcounter{section}{0}
\section{Algorithm Details}
\label{sup:proof}

This section elaborates on the specific implementation details of the four core algorithms mentioned in the main text. These include three optimization algorithms for performance comparison: Beam Search Assembly Optimization (Algorithm \ref{alg:beam_only}), Genetic Assembly Optimization (Algorithm \ref{alg:ga_only}), and Langevin-like Updates for Beam Search Assembly Optimization (Algorithm \ref{alg:beam_langevin}), along with the Sinkhorn algorithm (Algorithm \ref{alg:sinkhorn}) referenced in Section 3.2 of the main text. The Sinkhorn algorithm serves to convert the discrete selection matrix into a continuous representation, providing the foundation for continuous Langevin sampling optimization. It is noteworthy that the Sinkhorn algorithm represents a conventional approach used solely for comparison and is not employed in this work.

\renewcommand{\thealgorithm}{S1}
\begin{algorithm}[]
	\footnotesize
	\caption{Beam Search Assembly Optimization (BS-only)}
	\label{alg:beam_only}
	
	\KwIn{
		A set of video shots $\mathbf{V}=\{v_1,\ldots,v_N\}$;
		Sequence length $M$; Maximum number of iterations $I$;
		Number of desired top solutions $Q$;
		Energy function $E$;
		Beam size $B$.
	}
	\KwOut{A set of top video sequences and their corresponding scores}
	
	\BlankLine
	\textbf{Initialize}: Randomly generate $B$ candidate sequences $\mathcal{B}=\{S_1,\ldots,S_B\}$; set the best energy $E_{\text{best}} \leftarrow \infty$.\\
	
	\For{$iter=1$ \KwTo $I$}{
		Compute the energy $\{E(S) : S \in \mathcal{B}\}$ for all sequences in the current beam.
		
		\ForEach{$S \in \mathcal{B}$}{
			\If{$E(S)<E_{\text{best}}$}{
				Update $E_{\text{best}}$ and record the current best solution. If there are multiple equally good solutions, record all of them.
			}
		}
		
		\If{\text{enough high-quality solutions ($>Q$) have been found}}{
			\textbf{break}
		}
		
		\tcp{Beam Expansion}
		$\mathcal{C} \leftarrow \emptyset$; \\
		\ForEach{$S \in \mathcal{B}$}{
			Generate a neighborhood set $\mathcal{N}(S)$ by local operations such as swapping or replacing elements to create candidate neighbors;\\
			Insert $(S, E(S))$ into $\mathcal{C}$;\\
			For each neighbor $S' \in \mathcal{N}(S)$, compute $E(S')$ and insert $(S', E(S'))$ into $\mathcal{C}$.
		}
		
		Sort $\mathcal{C}$ by energy in ascending order. Select the top $B$ solutions to form the new beam: $\mathcal{B} \leftarrow \text{Top-}B(\mathcal{C})$.
	}
	
	\Return{The list of best sequences and their corresponding scores.}
	
\end{algorithm}

\renewcommand{\thealgorithm}{S2}
\begin{algorithm}[]
	\footnotesize	
	\caption{Genetic Assembly Optimization (GA-only)}
	\label{alg:ga_only}

	\KwIn{
		A set of video shots $\mathbf{V}=\{v_1,\ldots,v_N\}$;
		Sequence length $M$; Maximum number of iterations $I$;
		Number of desired top solutions $Q$;
		Energy function $E$;
		GA parameters: crossover probability $r_c$, mutation probability $r_m$, population size $P$.
	}
	
	\KwOut{
		A set of best sequences and their corresponding scores.
	}
	
	\BlankLine
	\textbf{Initialize}: Randomly generate a population of $P$ individuals $\mathcal{P} = \{\mathbf{S}_1, \ldots, \mathbf{S}_P\}$; set $E_{\text{best}} \leftarrow \infty$.\\
	
	\For{$iter=1$ \KwTo $I$}{
		\textbf{Fitness Evaluation}:Compute $F(\mathbf{S}) = -E(\mathbf{S})$ for each $\mathbf{S} \in \mathcal{P}$.\\
		Find $E_{\min}$ in the population. If $E_{\min} < E_{\text{best}}$, update the best solution. For ties, record all optimal solutions.\\
		\textbf{Selection Probability}: Compute the selection probability for each sequence $\mathbf{S}$:
		\[
		p(\mathbf{S}) = \frac{\exp(F(\mathbf{S}))}{\sum_{\mathbf{S}' \in \mathcal{P}} \exp(F(\mathbf{S}'))}.
		\]
		Sample $P$ sequences from $\mathcal{P}$ according to $p(\mathbf{S})$ to form the parent candidate set $\mathcal{P}''$.\\
		
		\tcp{Crossover \& Mutation}
		$\mathcal{P}_{new}=\emptyset$;\\
		\While{$|\mathcal{P}_{new}|<P$}{
			Randomly select two parents $S_{p1}, S_{p2}\in\mathcal{P}''$;\\
			With probability $r_c$, perform crossover to obtain offspring $(C_1,C_2)$; otherwise $(C_1,C_2)\leftarrow (S_{p1},S_{p2})$;\\
			With probability $r_m$, mutate $C_1$ and $C_2$;\\
			Insert the offspring into $\mathcal{P}_{new}$.
		}
		$\mathcal{P}\leftarrow\mathcal{P}_{new}$;\\
		\If{\text{enough high-quality solutions ($>Q$) are found}}{
			\textbf{break}
		}
	}
	\Return{The top solutions and their corresponding scores.}
\end{algorithm}

\renewcommand{\thealgorithm}{S3}
\begin{algorithm}[]
	\footnotesize
	\caption{Langevin-like Updates for Beam Search Assembly Optimization (Langevin+BS)}
	\label{alg:beam_langevin}
	\KwIn{
		A set of video shots $\mathbf{V}=\{v_1,\ldots,v_N\}$;
		Sequence length $M$; Maximum number of iterations $I$;
		Number of desired top solutions $Q$;
		Energy function $E$;
		Langevin parameters: step size $\epsilon$, temperature $T$;
		Beam size $B$.
	}
	
	\KwOut{Top sequences and their corresponding scores}
	
	\BlankLine
	\textbf{Initialize}: Randomly generate $B$ sequences $\mathcal{B}=\{S_1,\ldots,S_B\}$; set $E_{\text{best}} \leftarrow \infty$.\\
	\For{$iter=1$ \KwTo $I$}{
		\tcp{Langevin-like Local Optimization}
		$\mathcal{B}' \leftarrow \emptyset$; \\
		\ForEach{$S \in \mathcal{B}$}{
			Compute $E(S)$; \\
			Generate the neighborhood $\mathcal{N}(S)$ by operations like swapping or replacing; \\
			Find the neighbor with the lowest energy: $S_{\text{best\_neighbor}} = \arg\min_{S' \in \mathcal{N}(S)} E(S')$;\\
			Compute $\Delta E = E(S_{\text{best\_neighbor}})-E(S)$; \\
			Compute acceptance probability $P_{\text{accept}} = \min(1, \exp(-\Delta E / (\epsilon T)))$; \\
			\If{$\Delta E < 0$ or random number $< P_{\text{accept}}$}{
				$S \leftarrow S_{\text{best\_neighbor}}$
			}
			Insert the updated $S$ into $\mathcal{B}'$.
		}
		
		\tcp{Beam Search Operations}
		$\mathcal{B} \leftarrow \mathcal{B}'$; \\
		Compute energies for all sequences in $\mathcal{B}$ and update the global best solution set;\\
		\If{\text{enough high-quality solutions ($>Q$) are found}}{
			\textbf{break}
		}
		$\mathcal{C} \leftarrow \emptyset$; \\
		\ForEach{$S \in \mathcal{B}$}{
			Insert $(S,E(S))$ into $\mathcal{C}$; \\
			Generate neighbors $\mathcal{N}(S)$ again, compute $E(S')$ for $S' \in \mathcal{N}(S)$, and insert $(S',E(S'))$ into $\mathcal{C}$.
		}
		Sort $\mathcal{C}$ by ascending energy and select the top $B$: $\mathcal{B} \leftarrow \text{Top-}B(\mathcal{C})$;
	}
	\Return{Top sequences and corresponding scores}
\end{algorithm}

\renewcommand{\thealgorithm}{S4}
\begin{algorithm}[]
	\footnotesize
	\caption{Sinkhorn Algorithm}
	\label{alg:sinkhorn}
	\KwIn{Selection matrix \( \mathbf{X} \in \mathbb{R}^{N \times K} \), maximum iterations \( T \), hard assignment flag \(  \text{h} \)
	}
	\KwOut{Normalized matrix \( \mathbf{X}_{\text{s}} \in [0, 1]^{N \times K} \) or hard assignment matrix \( \mathbf{X}_{\text{h}} \)}
	\BlankLine
	Initialize log-normalized matrix: \( \mathbf{A} \gets \mathbf{X} \)	\\
	\For{\( t = 1, \dots, T \)}{
		\textbf{Column normalization}:
		\[
		\mathbf{A}_{ij} \gets \mathbf{A}_{ij} - \log \left( \sum_{i=1}^N \exp(\mathbf{A}_{ij}) \right), \quad \forall j \in \{1, \dots, K\}
		\] \\[4pt]
		\textbf{Row normalization}:
		\[
		\mathbf{A}_{ij} \gets \mathbf{A}_{ij} - \log \left( \sum_{j=1}^K \exp(\mathbf{A}_{ij}) \right), \quad \forall i \in \{1, \dots, N\}
		\]
	}
	Compute normalized matrix: \( \mathbf{X}_{\text{s}} \gets \exp(\mathbf{A}) \)\\[4pt]
	\If{\( \text{h} = \text{True} \)}{
		Determine the index of the maximum element in each column:
		\[
		i_k = \arg\max_{i} X_{ik}, \quad \forall k \in \{1, \dots, K\}
		\]
		Construct the hard assignment matrix:
		\[
		X_{ik}^{\text{h}} =
		\begin{cases} 
			1, & \text{if } i = i_k \\[3pt]
			0, & \text{otherwise}
		\end{cases}
		\]\\[4pt]
		\Return{Hard assignment matrix \( \mathbf{X}_{\text{h}} \)} 
	}
	\Return{Normalized matrix \( \mathbf{X}_{\text{s}}  \)} 
\end{algorithm}

\renewcommand\thesection{\Alph{section}}
\setcounter{section}{2}
\section{Optimization process.}
\label{sup:opt_pro}
To provide a more comprehensive demonstration of our adopted optimization method, this section showcases the optimization process of jointly optimizing shot size and motion syntaxes using discrete Langevin-based sampling in the “5 choose 3” and “10 choose 7” scenarios.

Figure \ref{fig:opt_pro}(a) illustrates the process of jointly optimizing the shot-scale and motion syntaxes for the “5 choose 3” scenario using discrete Langevin-based sampling. In this scenario, the theoretical maximum score for optimizing the shot-scale syntax and the motion syntax individually is 2. For the joint optimization, we take half of each respective energy function, which results in a theoretical maximum score of 2 for the joint task as well. As the figure shows, our method consistently finds the optimal combination in fewer than 100 iterations.

Figure \ref{fig:opt_pro}(b) illustrates the process of jointly optimizing the shot-scale and motion syntaxes for the “10 choose 7” scenario using discrete Langevin-based sampling. In this scenario, as the energy score is the negative of the best syntax score, the optimal joint syntax score for the selected video sequence arrangement is 6. Similarly, our method finds the optimal combination in fewer than 100 iterations.

In summary, our method, which simultaneously optimizes both shot size and motion syntaxes, is capable of finding the optimal combination within a small number of iterations in both scenarios, thus demonstrating its effectiveness. The approach effectively explores the solution space to identify high-quality video sequences with low energy, achieves strong performance in complex scenarios, and exhibits good scalability.

\renewcommand{\thefigure}{S1}
\begin{figure}[]
		\centering
		\includegraphics[width=0.8\columnwidth]{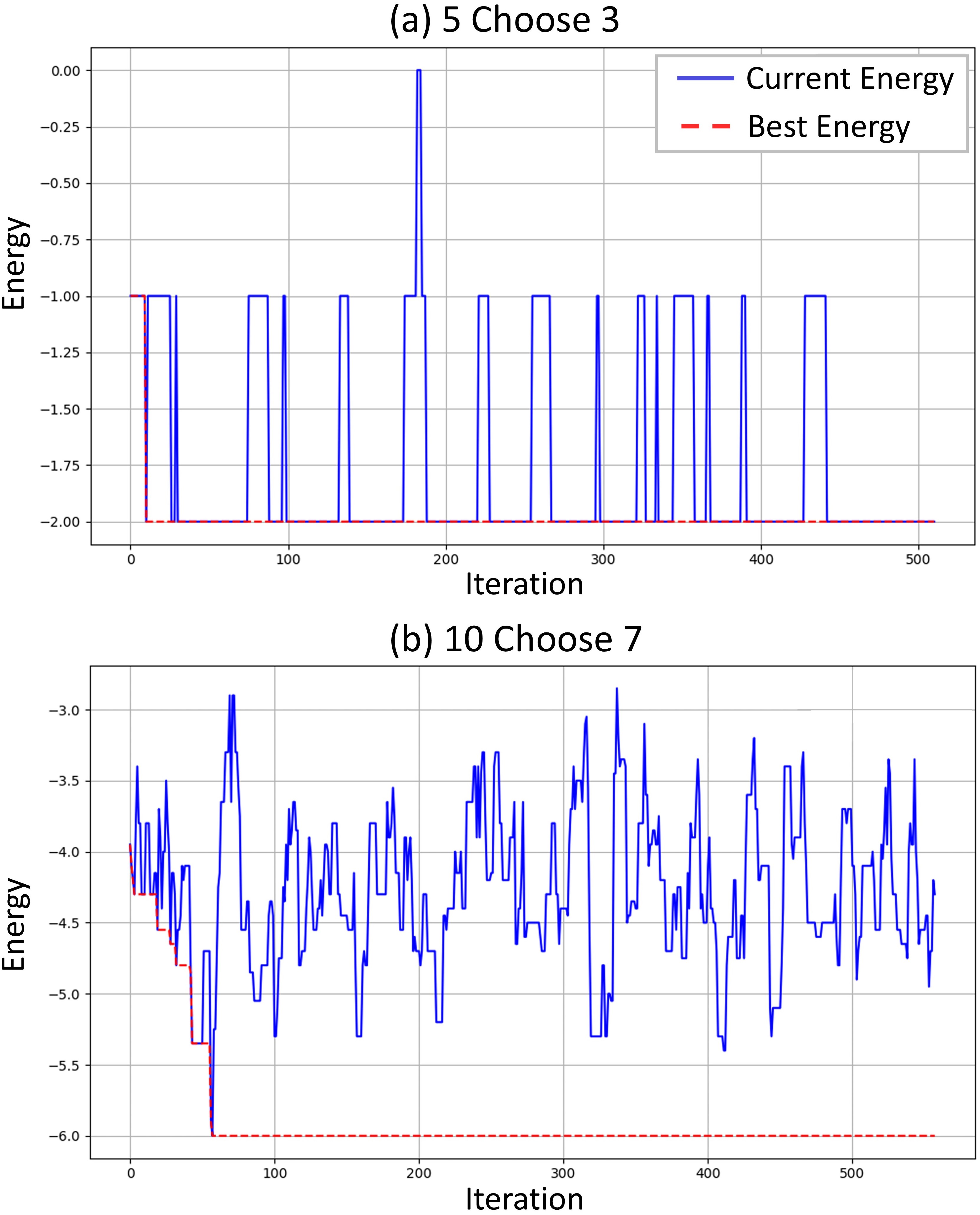}
		\caption{Langevin Sampling Video Assembly Optimization Process.}
		\label{fig:opt_pro}
	\end{figure}
	
\end{document}